\documentclass[lettersize,journal]{IEEEtran}
\usepackage{amsmath,amsfonts}
\usepackage{algorithmic}
\usepackage{algorithm}
\usepackage{array}
\usepackage[caption=false,font=normalsize,labelfont=sf,textfont=sf]{subfig}
\usepackage{textcomp}
\usepackage{stfloats}
\usepackage{url}
\usepackage{verbatim}
\usepackage{graphicx}
\usepackage{color}
\usepackage{cite}
\usepackage{makecell}
\begin{document}

\title{ScribbleSense: Generative Scribble-Based Texture Editing with Intent Prediction}

\author{Yudi Zhang, Yeming Geng, and Lei Zhang,~\IEEEmembership{Member,~IEEE}
\thanks{Yudi Zhang, Yeming Geng and Lei Zhang are with the School of Computer Science,
Beijing Institute of Technology, Beijing 100081, China.}
\thanks{Manuscript received xxxx xx, 2025; revised xxxx xx, 2025.}}

\markboth{Journal of \LaTeX\ Class Files,~Vol.~xx, No.~xx, July~2025}%
{Shell \MakeLowercase{\textit{et al.}}: ScribbleSense: Generative Scribble-Based Texture Editing with Intent Prediction}


\maketitle

\begin{abstract}
Interactive 3D model texture editing presents enhanced opportunities for creating 3D assets, with freehand drawing style offering the most intuitive experience. However, existing methods primarily support sketch-based interactions for outlining, while the utilization of coarse-grained scribble-based interaction remains limited. Furthermore, current methodologies often encounter challenges due to the abstract nature of scribble instructions, which can result in ambiguous editing intentions and unclear target semantic locations. To address these issues, we propose ScribbleSense, an editing method that combines multimodal large language models (MLLMs) and image generation models to effectively resolve these challenges. We leverage the visual capabilities of MLLMs to predict the editing intent behind the scribbles. Once the semantic intent of the scribble is discerned, we employ globally generated images to extract local texture details, thereby anchoring local semantics and alleviating ambiguities concerning the target semantic locations. Experimental results indicate that our method effectively leverages the strengths of MLLMs, achieving state-of-the-art interactive editing performance for scribble-based texture editing.
\end{abstract}


\section{Introduction}
\label{sec:intro}
\IEEEPARstart{T}{he} advancement of generative artificial intelligence has led to the increasing application of generative models in texture editing tasks, with the objective of producing high-quality textures that are consistent with real-world image distributions through interactive methods. Current texture editing methods utilize various types of inputs, including text, dragging, and freehand scribbles. Among these, scribble-based editing allows users to express their editing intentions in a straightforward and intuitive manner. Users can apply scribbles of the desired color at target locations to generate the corresponding semantic texture details. This straightforward color block scribbling approach minimizes user input requirements, eliminates the complexity associated with crafting high-quality prompts, and facilitates precise control over the color of local textures. Consequently, this editing method has the potential for widespread applications in fields such as virtual reality, computer games and other digital entertainment.

In contrast to text-based \cite{guerrero2024texsliders,youwang2024paint,dong2024coin3d}, image-based \cite{yeh2024texturedreamer,xie2024styletex}, or sketch-based \cite{mikaeili2023sked,liu2024sketchdream} texture editing methods, research on scribble-based texture editing remains limited. Notable works in this area include TEXTure \cite{richardson2023texture}, which utilizes existing inpainting models to fill designated areas with textures that correspond to the colors of scribbles. Although the application of diffusion models enhances the plausibility of the generated textures, the results frequently misalign with user intentions. This issue can be attributed to three primary factors. First, the framework lacks prior knowledge of user behavior, hindering its capacity to accurately interpret editing intentions. Second, even with additional textual prompts, the tension between global text descriptions and local scribble editing targets creates semantic ambiguity in spatial positioning, leading to unsatisfactory edits. Finally, the limited diversity in the training data of image generation models restricts the model's capacity for ``imagination'', which results in failures in certain editing scenarios.

\begin{figure*}[htbp]
  \centering
   \includegraphics[width=0.97\linewidth]{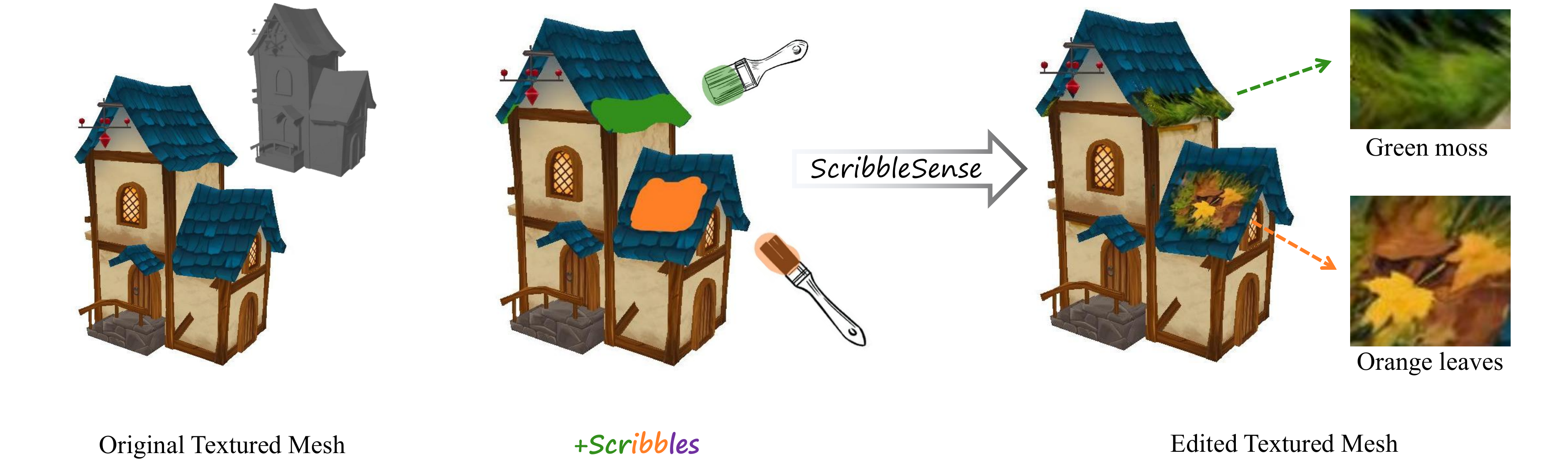}
   \caption{By freely scribbling with different colors on the surface of the 3D model, our model can reasonably predict the user's editing intentions and generate high-quality local textures that match the scribble colors.}
   \label{fig:teaser}
\end{figure*}

To address the issue, we first note that LLMs possess extensive commonsense knowledge and the ability to comprehend and learn from user behavior. These attributes can effectively mitigate the limitations associated with using only image diffusion models, thereby enhancing the understanding of editing intentions. Furthermore, we contend that transforming the global image conditional control editing task into a local generation task addresses both semantic spatial ambiguity and the challenge of limited global image diversity in generation. Specifically, utilizing the local semantic control predicted by MLLMs to produce local textures resolves semantic ambiguity at each positional context. Moreover, by incorporating the geometric information of the 3D mesh, we further refine the editing mask to better align with the user's true intent. Additionally, as local textures can be derived from a diverse array of complete images, this approach reduces the domain gap between the target generation content and the training data. Ultimately, integrating the diffusion model facilitates the seamless incorporation of local textures into global textures, yielding results that align with the intended editing objectives (see Figure \ref{fig:teaser}). 

The main contribution of our work is the development of a novel scribble-based texture editing framework. Specifically, we are the first to apply multimodal LLMs to texture editing tasks by predicting editing intentions. We evaluate the adaptability of MLLMs to this task from various perspectives, enabling burden-free scribble editing without the need for annotations. To address the common issue of local edit failures caused by semantic spatial ambiguity in global generation methods, we introduce a pipeline that generates and integrates contextually appropriate local textures. Additionally, we leverage geometry information from the 3D mesh to refine the scribbled area mask, which improves the robustness of our method when dealing with imprecise or coarse inputs. We demonstrate the effectiveness of our method through a series of experiments and further explore a variety of interaction methods and stylized generation techniques within our original framework, showcasing the scalability of our method.

\section{Related Work}
\label{sec:relatedwork}

\subsection{Generative Texture Editing Methods}
Generative texture editing methods can be broadly divided into global and local editing. For a given textured mesh, several text-based approaches \cite{metzer2023latent,chen2023text2tex,cao2023texfusion,zeng2024paint3d,youwang2024paint,liu2024text} regenerate textures from target text prompts, thus enabling global edits. However, these methods overlook the original texture as a reference, making it difficult to apply fine-grained modifications while retaining existing details. To overcome this issue, ITEM3D \cite{liu2024directional} defines a relative direction between the source and target text prompts to guide the optimization of the original texture, achieving more precise edits. Similarly, Instructive3D \cite{kathare2025instructive3d} uses a tri-plane representation and incorporates text embeddings into the denoising process to edit 3D models effectively.

For the local texture editing, approaches typically involve inputting a local mask and editing prompt. 3D Paintbrush \cite{decatur20243d} uses Cascade Score Distillation, allowing the distillation of scores across multiple resolutions, enabling high-quality generation of local textures. TEXGen \cite{yu2024texgen} fine-tunes diffusion models with texture datasets to support texture inpainting.

Compared to text-based local editing, drawing-based editing offers a more intuitive and interactive experience. SKED \cite{mikaeili2023sked} and SketchDream \cite{liu2024sketchdream}, built on NeRF \cite{mildenhall2021nerf} representations, enable local sketch-based editing but typically alter both geometry and texture simultaneously. TEXTure \cite{richardson2023texture} accepts colored scribbles as input and applies a checkerboard-style mask within the scribbled region to guide texture generation via an image diffusion model. However, its inpainting process often blends the scribble color with the original texture, resulting in unintended outcomes. Alternatively, Diffusion Texture Painting \cite{hu2024diffusion} uses images as ``brushes'', supporting local texture drawing. However, this method entails considerable time investment in selecting painting images, which diminishes the advantages of scribble-based editing. In light of this, our approach provides a more intuitive and user-friendly texture editing experience that better aligns with user intent, thus addressing a significant gap in this area.

\subsection{LLMs in Visual Content Generation Tasks}
Large language models (LLMs) such as ChatGPT \cite{brown2020language} and Llama \cite{touvron2023llama} have made rapid progress. A large number of visual generation tasks are controlled by text-based instructions, making prompt engineering the most direct bridge for leveraging LLMs in text-to-image and text-to-3D tasks. For instance, Promptify \cite{brade2023promptify} enhances image quality by optimizing prompts via LLMs, while PromptMagician \cite{feng2023promptmagician}, Promptmaker \cite{jiang2022promptmaker}, and Reprompt \cite{wang2023reprompt} achieve similar improvements. Beyond prompt optimization, some works employ LLMs to generate parameters for 3D models. 3D-GPT \cite{sun20233d} translates user instructions into Blender parameters for geometry and texture editing. ELLA \cite{hu2024ella} uses LLMs as text encoders for high-quality embeddings, and DiffusionGPT \cite{qin2024diffusiongpt} and GenArtist \cite{wang2025genartist} act as agent systems to select expert models for better generation outcomes.

Multimodal LLMs, especially Visual Language Models (VLMs) such as GPT-4 \cite{achiam2023gpt}, Gemini \cite{team2024gemini}, and InternVL \cite{zhu2025internvl3}, further expand possibilities for visual content generation. BlenderAlchemy \cite{huang2024blenderalchemy} integrates VLMs for texture editing in Blender, and CompAgent \cite{wang2024divide} parses spatial relations in prompts and produces layout images to assist with image generation. MagicQuill \cite{liu2025magicquill} leverages multimodal LLMs to interpret contour sketches and color edits, enabling intelligent image editing.

VLMs have a wide range of applications in the domain of visual perception, including image captioning \cite{xu2024pixel}, optical character recognition (OCR) \cite{luo2024layoutllm,ye2023ureader}, image segmentation \cite{lai2024lisa,xia2024gsva}, and visual grounding \cite{wan2024contrastive,jiang2024visual}. However, these abilities are seldom linked to visual generation. In our task, the model must both understand visual inputs and use this semantic knowledge to generate new content. Consequently, the incorporation of MLLMs is the most suitable choice for this task, and our task, in turn, expands the application scope of MLLMs.

\section{Method}

\begin{figure*}[htbp]
  \centering
   \includegraphics[width=0.95\linewidth]{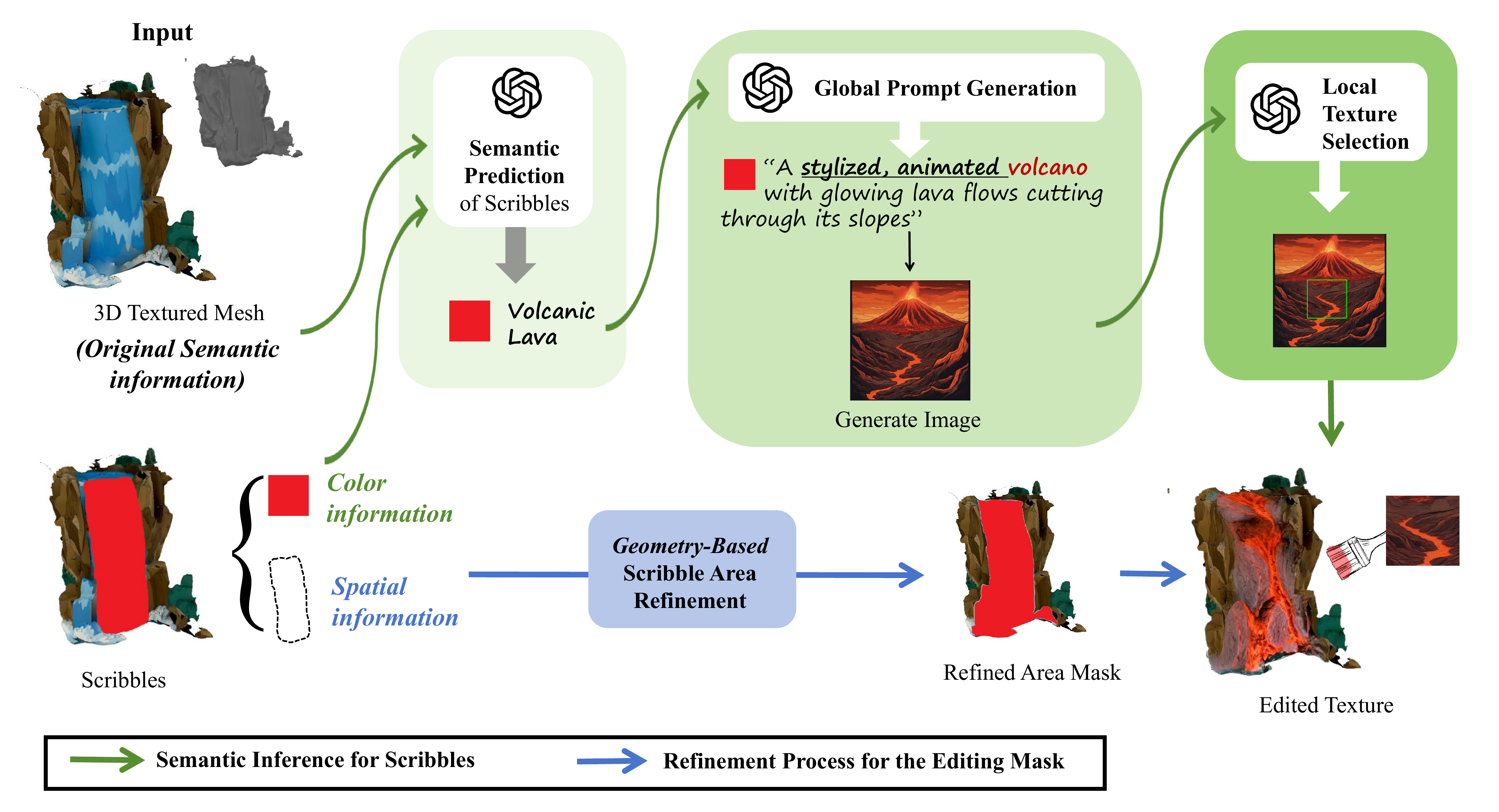}
   \caption{Overview of ScribbleSense. We leverage multimodal LLMs in our pipeline for semantic prediction, global prompt generation, and local texture selection for texture editing.}
   \label{fig:pipeline}
\end{figure*}

Figure \ref{fig:pipeline} illustrates an overview of our method. The inputs are the 3D textured mesh and the user-scribbled images. We decompose the scribble into two components: color information and spatial information, which are used in the subsequent semantic reasoning and mask refinement processes. For semantic inference, we first input the multi-view rendered images of the original texture along with the scribbled images into a multimodal LLM to obtain a basic semantic prediction of the scribble. Based on this prediction, we use the LLM to generate a refined global prompt, allowing us to produce a high-quality global image with the intended local texture. The LLM is then used to select the most relevant texture patch. For mask refinement, we incorporate the geometry of the 3D model to further localize and refine the scribbled area, available as an optional module. Finally, the selected local texture is integrated into the refined area using an inpainting model.

\subsection{Prediction of Editing Intent}

For highly abstract or unclear scribble instructions, compared to image diffusion models that rely solely on visual priors,  LLMs can incorporate richer semantic prior knowledge, allowing them to capture the user's editing intent more accurately. Therefore, we introduce multimodal LLM into our pipeline for editing intent prediction.

Furthermore, we found that multi-view images are essential for predicting editing intent. Multi-view images provide more comprehensive semantic and appearance information, helping to improve the accuracy of predictions and the consistency of editing results.

{\bf Semantic Information.} 
A single scribble image may lead to misinterpretations due to insufficient information. For example, a red scribble on a mountain seen from one viewpoint might be interpreted by the MLLM as lava. However, when multi-view original images are provided, the MLLM receives more contextual information. If an image from another viewpoint shows snow on the mountain, the MLLM can interpret the same red scribble as iron-rich rocks rather than lava. This inference arises from the overall consistency of the scene.

{\bf Appearance Information.} From a single viewpoint, when a global reference is absent, the model may produce textures that randomly appear for the predicted semantics, which may not align with the existing scene. However, if other viewpoints feature textures that share the same semantics, the model can utilize these existing textures to create new ones that exhibit a consistent aesthetic style. This enhances the overall coherence and realism of the scene.

To fully leverage multi-view information, we select four rendered images of a 3D model with original textures, taken from $\theta = 0^{\circ}$, $\phi = \{0^{\circ}$, $90^{\circ}$, $180^{\circ}$, $270^{\circ}$\}. These images, along with the scribble content, are input into an MLLM, which is then tasked with predicting our editing intent. Figure \ref{fig:chat}.(a) demonstrates an example of the prompt and the MLLM's performance. With the multi-view images as input, we ask the model to comprehensively integrate the information from all 4 images to predict the most likely target semantics that align with the scribble color. These predicted results will serve as a reference for subsequent local texture generation, ensuring that the texture is highly consistent with the user's editing intent.

\subsection{Generating Consistent Local Textures with Predicted Semantics}

After obtaining the local semantics from the scribbling, inspired by Diffusion Texture Painting \cite{hu2024diffusion}, we first generate local textures representing the scribble semantics as ``brushstrokes'', which serve as the basic elements for editing. Using text-to-image generation, we can create local textures corresponding to the semantics. However, the directly generated image results are often unpredictable, and it is challenging to find optimal prompts to generate the desired image. To avoid this time-consuming trial-and-error process, we use LLM to generate detailed prompts that specify the desired content, including the scribble color, semantics, and target style, thereby improving the probability of generating high-quality local textures.

\begin{figure}[htbp]
  \centering
   \includegraphics[width=0.98\linewidth]{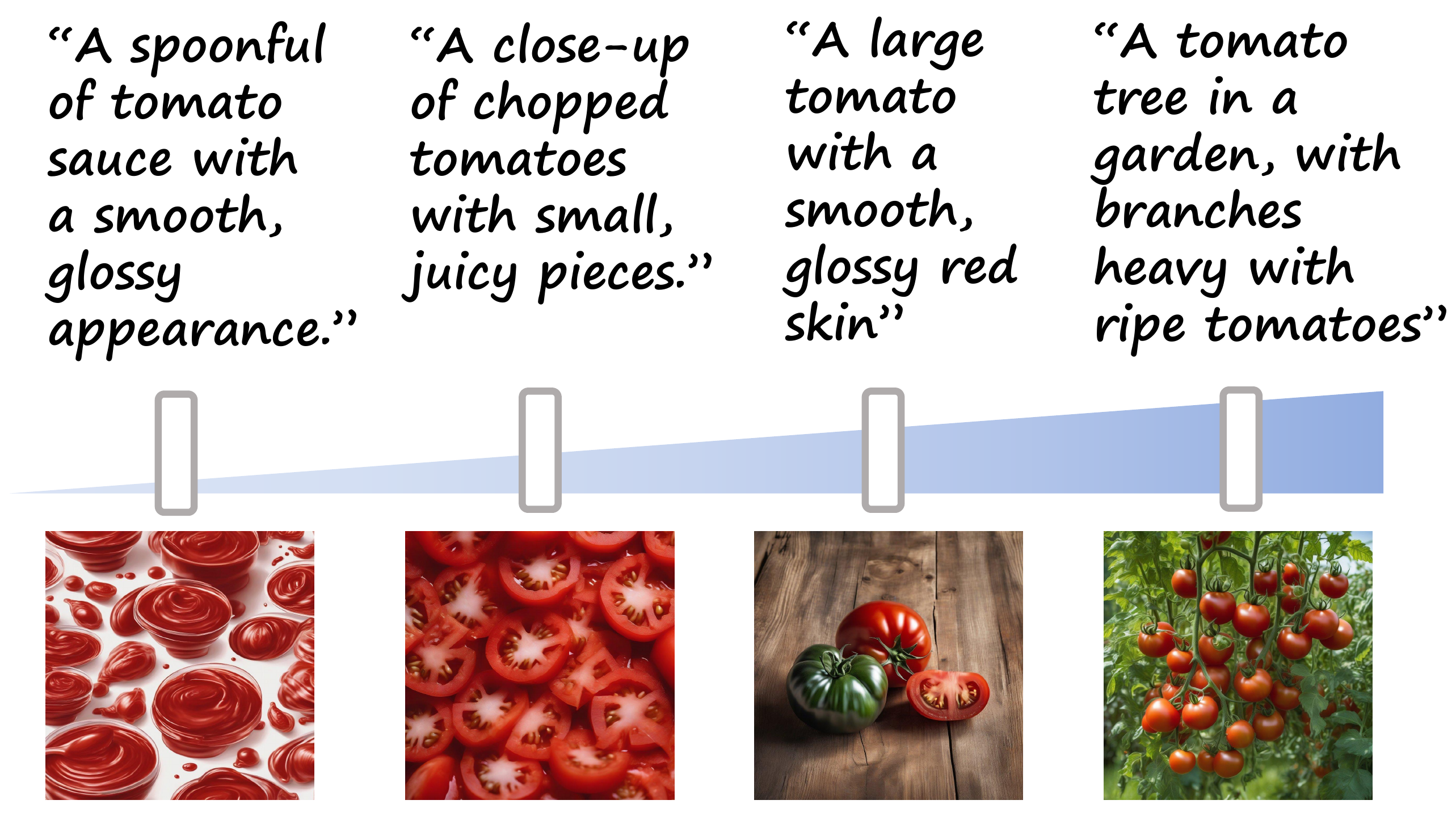}

   \caption{For the same local predicted semantics, using different global prompts will generate different forms of the object.}
   \label{fig:global}
\end{figure}

\begin{figure}[htbp]
  \centering
   \includegraphics[width=0.98\linewidth]{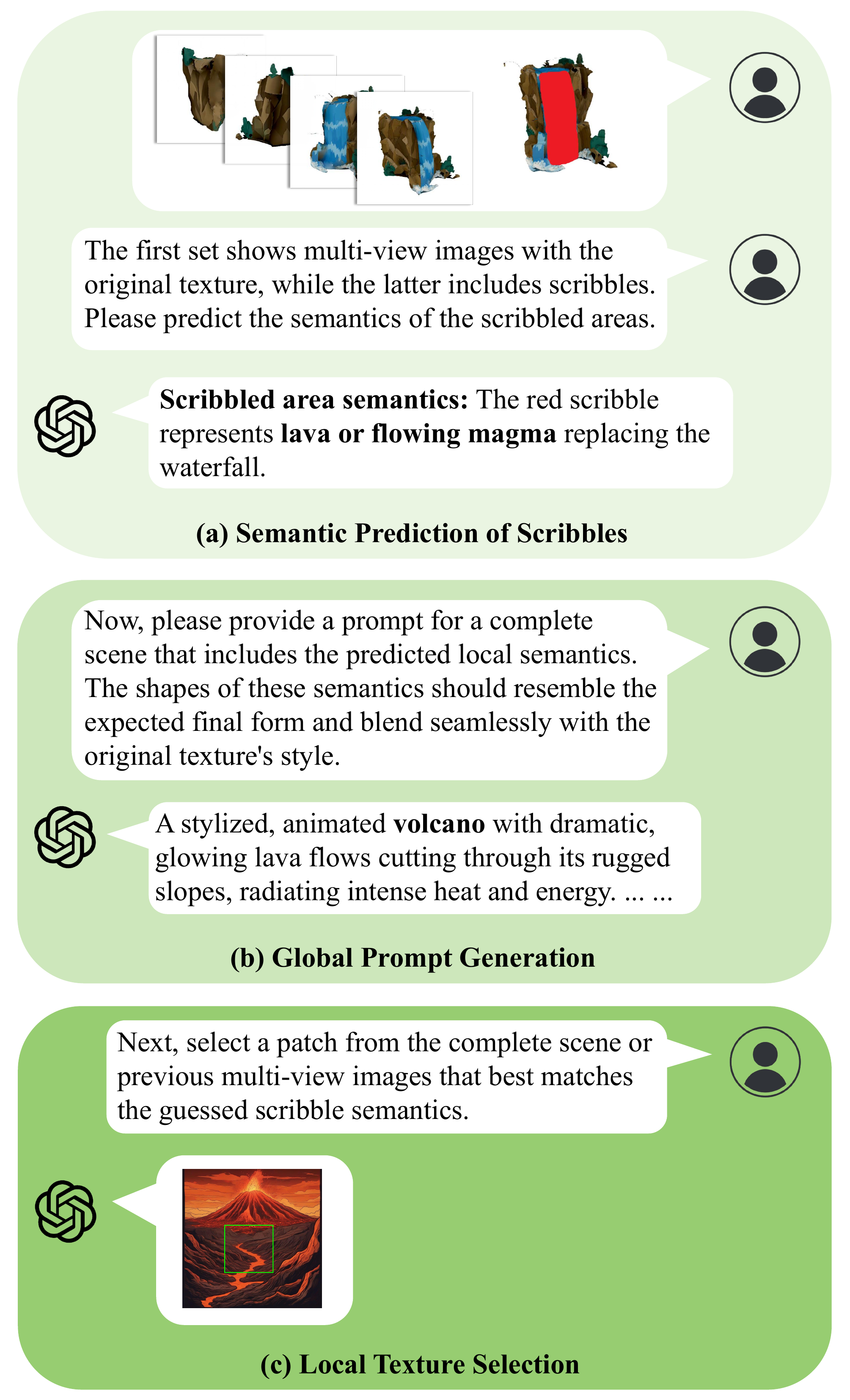}

   \caption{Examples of dialogues with the LLM at different stages of the pipeline.}
   \label{fig:chat}
\end{figure}

Additionally, we acknowledge a further challenge in the direct generation of local textures. The training datasets for diffusion models predominantly consist of images depicting whole objects, rather than emphasizing detailed components. Meanwhile, as shown in Figure \ref{fig:global}, for the same local semantics, using different global prompts can generate vastly different forms of the same semantics. Therefore, for a given scribble input with specified conditions, it is crucial to use a high-quality global prompt, rather than just a simple local description, to generate a locally textured form that fits the required conditions. Consequently, we prompt the LLM to produce a global scene description that encompass the local texture information. For instance, to generate lava, we request the LLM to create descriptive texts that include lava, such as ``volcano'', as illustrated in Figure \ref{fig:chat}.(b). Subsequently, we input these global scene descriptions into the latest Stable Diffusion model to generate several global images that integrate the local textures.

Upon generating global images with the requisite local textures, we proceed to identify the most suitable local texture from these images. Specifically, we leverage the MLLM to select the texture patch that best aligns with the local semantics, as shown in Figure \ref{fig:chat}.(c). 
Notably, when regions that closely correspond to the scribble semantics are identified within the original texture, we prioritize selecting patches from these regions to preserve the texture's stylistic consistency as much as possible. This strategy significantly enhances the coherence and realism of the generated image.

\subsection{Refining the Scribble Area Mask}
As previously discussed, the scribble input provided by the user is often abstract and imprecise, composed of two key elements: color and a spatial area (expressed as a mask). While color and shape can assist in semantic inference, the spatial area covered by the scribble often deviates from the user's actual editing intention due to its coarse and freehand nature. Therefore, in addition to semantic prediction, it is essential to refine the editing mask to better align its spatial extent with the user's intended editing area.

\begin{figure}[htbp]
  \centering
   \includegraphics[width=1.0\linewidth]{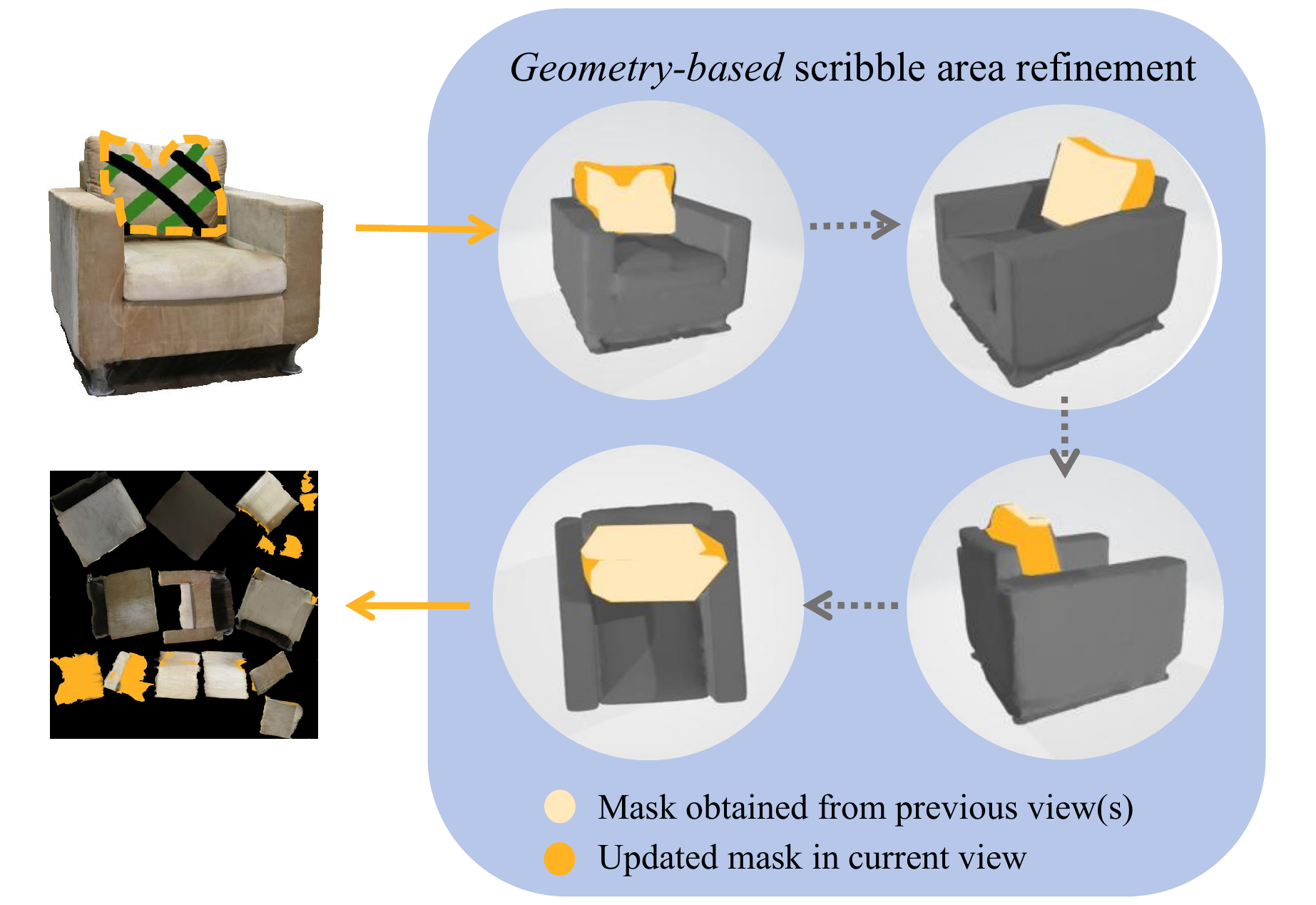}
   \caption{The rendering of textureless 3D meshes highlights the geometric structure of the model. To process the user's rough scribble, our method combines this geometric information with a segmentation model to iteratively refine the scribbled area mask.}
   \label{fig:geometry}
\end{figure}

To address this issue, we introduce a geometry-guided refinement module. Considering that the structure of textures is closely related to object geometry, we utilize multi-view renderings of the untextured mesh, and incorporate an image segmentation model to progressively improve the editing area. Specifically, in the initial scribble view, we apply the Segment Anything Model (SAM) \cite{kirillov2023segment} to obtain the minimal segmentation mask enclosing the scribbled input. Subsequently, in the next view, we map the previously refined mask area to the new viewpoint based on geometric correspondence, and apply SAM again to update the mask. This iterative refinement continues across all views. As shown in Figure \ref{fig:geometry}, the editing mask gradually evolves from a coarse sketch to a more precise spatial area, aligning more closely with the user's intent.

\subsection{Integrating Local Textures}

\begin{figure}[htbp]
  \centering
   \includegraphics[width=1.0\linewidth]{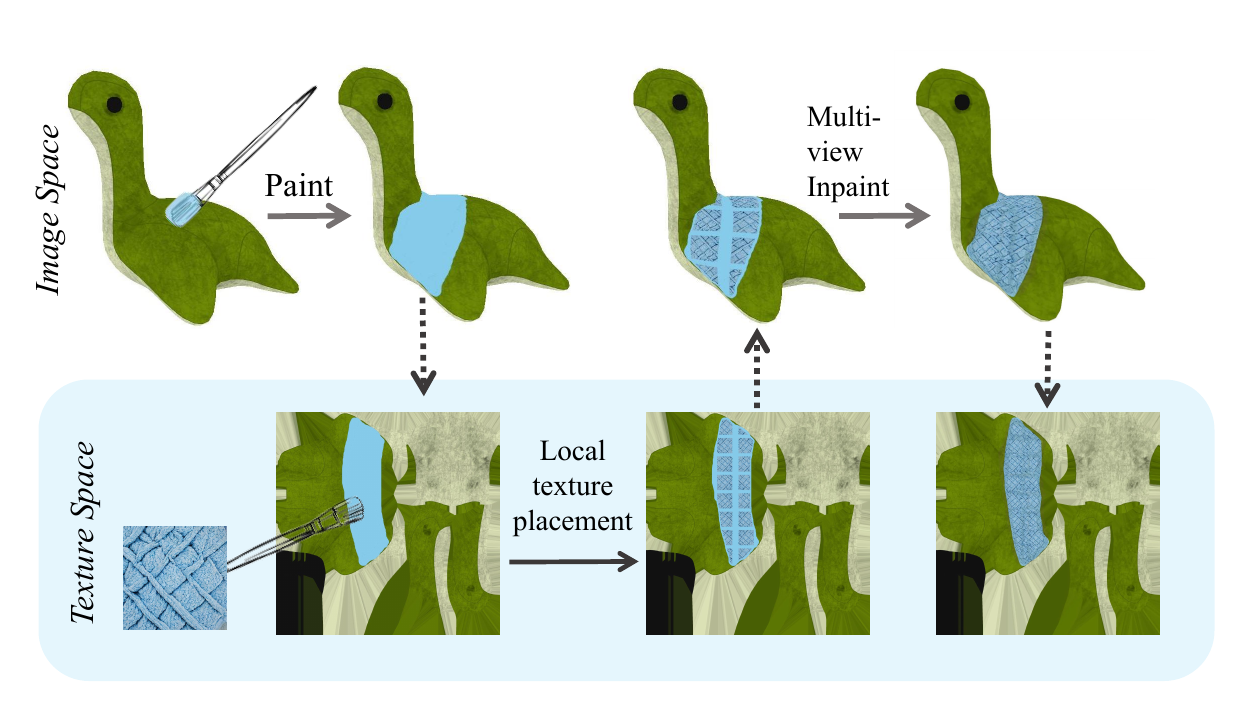}
   \caption{Pipeline of local texture integration. Texture patches are placed in the texture space, followed by inpainting in the image space.}
   \label{fig:tex_inpaint}
\end{figure}

To integrate the local texture patches into the refined scribble areas, we first determine the optimal fitting of the patches onto the 3D model surface. Once the patches are positioned, we employ an inpainting model to seamlessly blend any remaining unfilled regions. Figure \ref{fig:tex_inpaint} illustrates the process of texture patch placement and subsequent inpainting.

For patch placement, we adopt a straightforward and easily implementable subdivision method. Specifically, given an irregular scribble region $R$ in the texture space, we first compute its minimum bounding box, denoted as $B$. Let $B$ have a width $W$ and height $H$, while the local rectangular patches have a width $w$ and height $h$. The number of uniformly distributed local patches $P_{i,j}=(x_i,y_j)$ is given by:
\begin{equation}
    N=N_w\cdot N_h=\lfloor \frac{W}{w} \rfloor \cdot \lfloor \frac{H}{h} \rfloor
\end{equation}
The spacing between these patches is:
\begin{equation}
    \Delta x = \frac{W-N_w\cdot w}{N_w+1} ,
    \Delta y = \frac{H-N_h\cdot h}{N_h+1}
\end{equation}
Therefore, the position of each patch is determined by the following formula:
\begin{align}
\begin{cases}
        &x_i = i \cdot w + (i+1)\cdot \Delta x \\
        &y_j = j \cdot h + (j+1)\cdot \Delta y 
\end{cases}
\end{align}
After locating the local textures, patches satisfying $P_{i,j}\subseteq R$ are fully retained, while those meeting $P_{i,j}\cap R=\emptyset$ are entirely discarded. For cases where $P_{i,j}$ are partially within R, only the intersecting region $P_{i,j}\cap R$ is preserved.

This allows the local textures to be evenly distributed over the selected irregular region, effectively covering most of the scribbled area while minimizing empty spaces. During the blending process, we implement an erosion operation to prevent sharp seams caused by insufficient spacing between adjacent rectangular patches. This operation slightly contracts the patch boundaries inward, creating small gaps that facilitate the subsequent inpainting process and ensure a smooth transition of textures.

Finally, we use an inpainting model to iteratively fill the gaps across different viewpoints, seamlessly integrating the local textures and producing a natural completion of the scribbled region.

\section{Experiments}

In the experiment, we selected 3D textured meshes from various sources, including Sketchfab~\footnote{https://sketchfab.com} and Objaverse \cite{deitke2023objaverse}, covering multiple categories for testing. The experiments include 24 3D meshes and 32 groups of inputs. We also prepared the real editing intent of users for each scribble, which refers to the semantics of the scribble region, represented by a prompt. We chose GPT-4 as the default MLLM, and included additional tests using the latest open-source InternVL3 \cite{zhu2025internvl3} for comparison. We use SDXL \cite{podellsdxl} as the generative model for creating global images containing local textures, with a guidance scale set to 7.5. We also use the inpainting model from Stable Diffusion \cite{rombach2022high} to integrate local textures into the global image. During inpainting and mask refinement, we use 6 fixed side views plus top and bottom views for full model coverage. For semantic prediction, we generate 4 global prompts and select the patch whose average color best matches the scribble as the local texture. Our method requires approximately 16 seconds on average to complete a single editing process when running on an A100 GPU.

\subsection{Comparison with Baselines}
\label{sec:cpbaseline}

\begin{figure*}[htbp]
  \centering
   \includegraphics[width=1.0\linewidth]{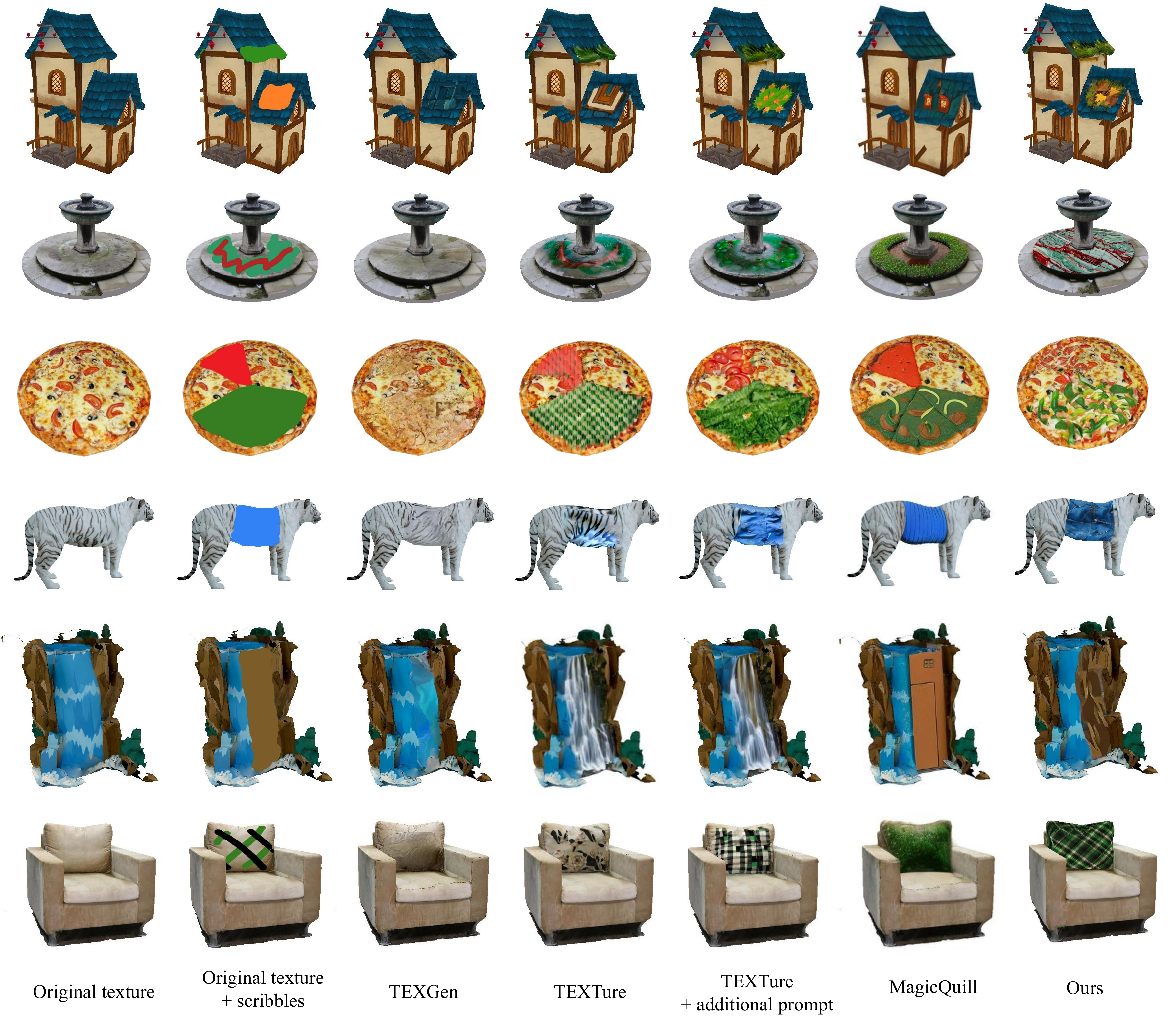}
   \caption{Qualitative comparison of our method with other baselines.}
   \label{fig:result}
\end{figure*}

In Section \ref{sec:relatedwork}, we mentioned that existing local editing methods can be divided into two categories. One common approach involves inputting a local mask of the editing region along with a prompt for editing, while the other approach is based on optional color scribble editing. Therefore, we selected one method from each category as the baseline for this part of the experiment. For the first category, we chose the most recent and representative TEXGen \cite{yu2024texgen}. For the second category, to the best of our knowledge, only the TEXTure \cite{richardson2023texture} framework currently supports this functionality. In addition, we also selected the current state-of-the-art MagicQuill \cite{liu2025magicquill} from 2D image editing methods that support scribble input, applying it to the views that contain the scribbles to enable a more comprehensive evaluation and comparison.


It is worth noting that our method, by leveraging MLLM predictions, does not necessitate the manual inclusion of prompts as editing references. In contrast, TEXTure requires textual assistance in addition to the input of scribbles in different colors. To facilitate a more precise comparison, we organized the TEXTure results into two distinct groups: one utilizing only the base prompt of the original 3D model, and the other incorporating an additional prompt derived from the real intent semantics of the scribble. As for TEXGen, which cannot specify scribble colors, we entered the real intent semantics to optimize its performance and ensure a fair comparison.

After selecting the baselines, we prepared multiple sets of models and scribble inputs for testing. Figure \ref{fig:result} shows the editing results obtained by different methods. From the results, it can be seen that even with related color hints in the text, TEXGen continues to experience editing failures due to the limitations inherent in its texture dataset. In the results for TEXTure without additional prompts, unreasonable editing outcomes are observed, primarily attributed to semantic ambiguity and confusion regarding editing intentions. Actually, in the absence of further instructions, the inpainting model forcibly merges the scribble color with the surrounding textures. In the group with additional prompt, the results are significantly improved, but there are still some issues with semantic localization being unclear and the quality of local scribbles being relatively low. The former issue is particularly evident in the waterfall example. Since ``rocks'' already appear in other parts of the texture, they are not accurately localized to the scribbled area during the overall texture generation process. Moreover, MagicQuill’s performance is unstable, producing unreasonable results in some cases. In contrast, our method effectively addresses these issues without relying on additional textual instructions, achieving texture editing that closely aligns with the intended editing objectives. A more detailed demonstration of our method's results is provided in the supplementary video.

\begin{figure*}[htbp]
  \centering
   \includegraphics[width=1.0\linewidth]{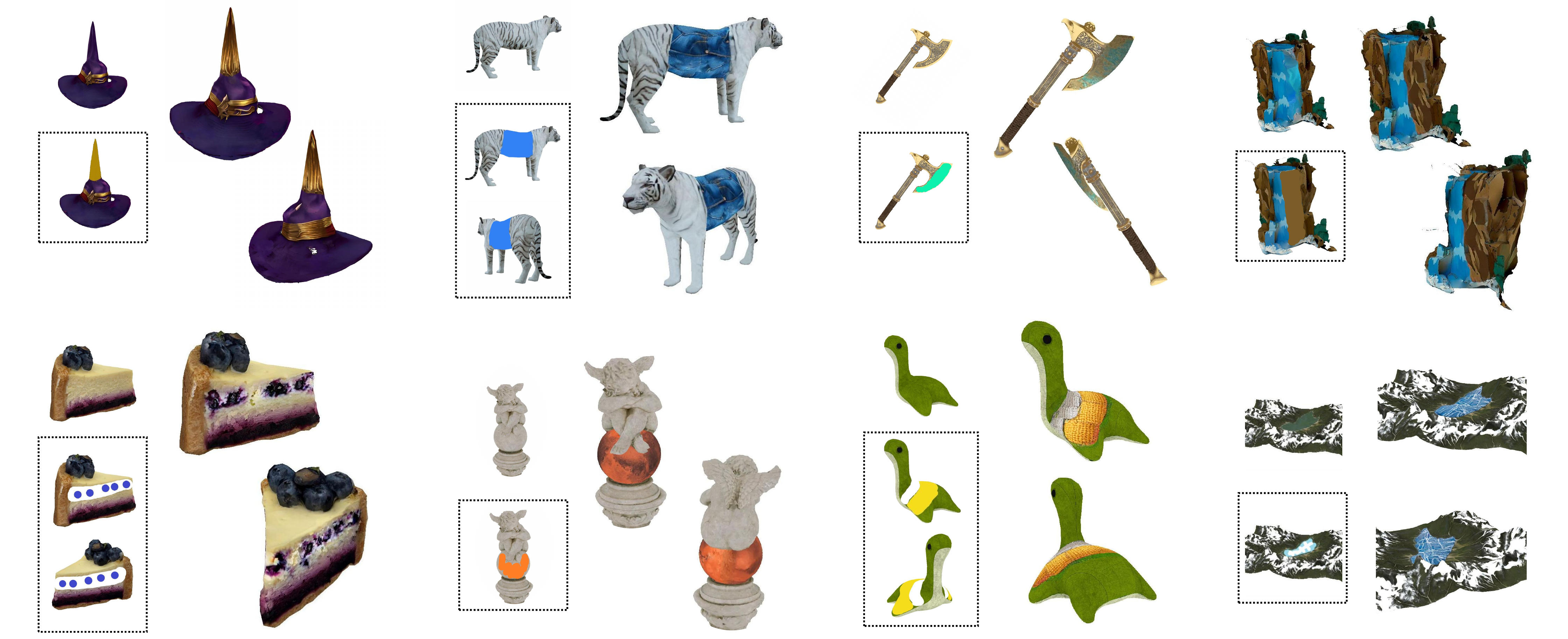}
   \caption{Multi-view results of our editing method on additional examples.}
   \label{fig:mvresults}
\end{figure*}
 
In the quantitative experiments, we first use single-view images containing the scribble region along with the prompt of real editing intent to calculate the Clip Score \cite{hessel2021clipscore} for each group's results, evaluating the quality of the final outcomes and their relevance to the editing intent. Using the model's original textures as a reference for realism, we also compare our method with the baselines using multi-view images and the FID \cite{heusel2017gans} metric. This allows us to further demonstrate the quality of the edits.
Furthermore, we assess user preferences for various editing methods through a user study, with 43 participants taking part in the survey. In addition to overall quality, we evaluate multi-view consistency using edited renderings from four evenly distributed viewpoints. As shown in Table \ref{tab:quantres}, our method performs comparably to the best existing approaches. Notably, as illustrated in Figure \ref{fig:result}, TEXGen often fails to apply edits, producing textures that remain close to the original ones, which may incidentally raise its consistency score. Figure \ref{fig:mvresults} further presents the multi-view editing results generated by our method. 

Since editing intent is subjective, we do not reveal the true intent of the scribble’s author in the user study. Instead, users choose the image that best matches the scribble's intent based on their understanding. Despite this, our method outperforms others in overall preference. This demonstrates that our approach not only adapts well to specific users but also has the ability to satisfy a broader range of users.
\begin{table}[!t]
\caption{Quantitative comparison with other baselines.\label{tab:quantres}}
\centering
\begin{tabular}{|c|c|c|c|c|}
\hline
Method & \makecell{CLIP-\\Score$\uparrow$} & FID$\downarrow$ & \makecell{User Preference\\ (Consistency) $\uparrow$} & \makecell{User Preference\\ (Overall)$\uparrow$}\\
\hline
TEXGen & 27.91 & 147.86 & {\bf 31.63 \%} & 13.49 \%\\
\hline
TEXTure & 29.12 & 161.87 & 9.30 \% & 14.88 \%\\
\hline
\makecell{TEXTure \\+ prompt} & 31.22 & 152.18 & 11.63 \% & 9.77 \%\\
\hline
MagicQuill & 28.04 & 150.12 & 16.74 \% & 17.21 \%\\
\hline
 Ours & {\bf 31.90} & {\bf 130.83} & \underline{30.70 \%} & {\bf 44.65 \%}\\
\hline
\end{tabular}
\end{table}

\subsection{Evaluation of MLLM}
\label{sec:evalllm}
{\bf Evaluation of Editing Intent Prediction Accuracy.} To further evaluate the compatibility of MLLM with this task, we assess its accuracy in predicting editing intent and compare it with existing works.  In the literature, the task of extracting semantics from a specified region in an input image is traditionally termed image captioning. 
Similarly, image recognition techniques capture the semantic information present throughout the entire image. Meanwhile, visual grounding tasks can identify semantic information in different regions of the image, collaborating with image recognition to achieve the functionality of image captioning.

Therefore, we select the following three methods as baselines for comparison in this part.
The first is the image dense caption feature from Microsoft’s Vision Studio, used for direct annotation.
We also select the representative image recognition model Recognize Anything Model (RAM) \cite{zhang2024recognize}, together with the image segmentation method Grounded SAM \cite{ren2024grounded}, for indirect image captioning.
We also include MagicQuill, which provides painting content prediction, to to enable a more thorough comparison.

As in Section \ref{sec:cpbaseline}, we selected multiple sets of models and scribble inputs, paired with the real editing intent of users as references. In addition, for the same model, we invited different users to create scribbles, thereby expanding the dataset used for this part of the evaluation. The evaluation criteria is that if keywords or synonyms from the real intent semantics appear in the annotated result for the scribble area, it is considered a correct prediction. Table \ref{tab:llmeval} presents a comparison of the prediction accuracy among the three baseline methods and the MLLM used in our approach.

Table \ref{tab:llmcontent} shows a comparison of the prediction results for two sets of inputs containing scribbles. It is evident that traditional methods, which are trained primarily on real-world images, lack a comprehensive understanding of scribbles. In both editing scenarios, they tend to focus on the shape and color of the scribbles, such as ``pink circle'', ``heart'', ``green'', rather than the semantics they convey. A similar issue can be observed in the results of MagicQuill, which is fine-tuned with edge maps incorporated and therefore focuses more on the semantics behind contours rather than those behind color regions. In contrast, our method, which leverages the common knowledge embedded in the MLLM and its understanding of user behavior, produces semantic predictions that closely align with the original texture. This capability accounts for the higher overall prediction accuracy of our method.

\begin{table}[!t]
\caption{Semantic prediction accuracy for scribbles across different models.\label{tab:llmeval}}
  \centering
  \begin{tabular}{|c|c|}
    \hline
    Method & \makecell{Accuracy of Editing\\ Intent Prediction} \\
    \hline
    Microsoft Vision Studio & 22.8 \% \\
    \hline
    RAM + GSAM & 13.5 \% \\
    \hline
    MagicQuill & 30.2 \% \\
    \hline
    MLLM (Ours) & {\bf 75.4 \%} \\
    \hline
    MLLM (Ours) with 4 predictions & {\bf 85.1 \%} \\
    \hline
    MLLM (Ours) w/o multi-view inputs & 71.6 \% \\
    \hline
  \end{tabular}
\end{table}

\begin{table}[!t]
\caption{Comparison of semantic predictions for scribbles across different models. \label{tab:llmcontent}}
  \centering
  \begin{tabular}{|c|c|c|}
    \hline
    Method & 
    \begin{minipage}[b]{0.28\columnwidth}
		\centering
		\raisebox{-.5\height}{\includegraphics[width=\linewidth]{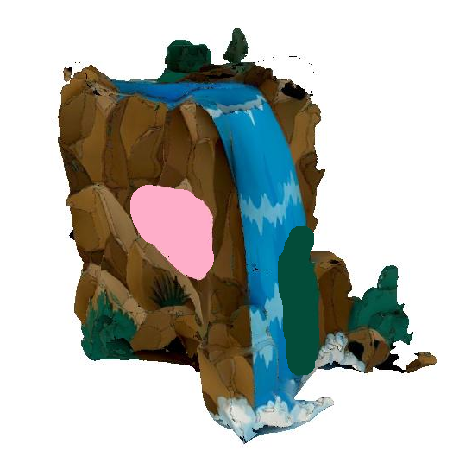}}
	\end{minipage} &
    \begin{minipage}[b]{0.28\columnwidth}
		\centering
		\raisebox{-.5\height}{\includegraphics[width=\linewidth]{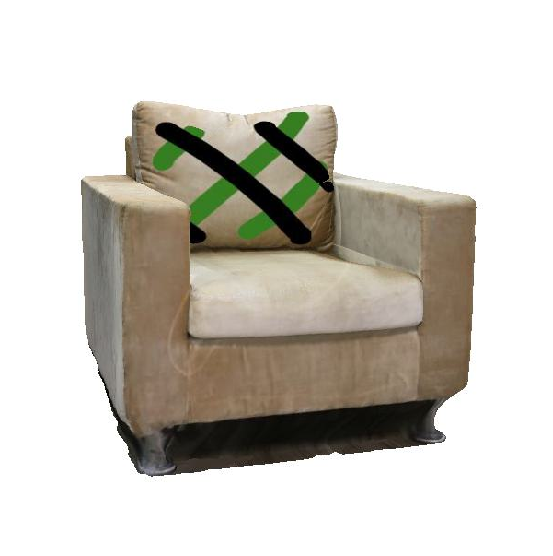}}
	\end{minipage}
    \\
    \hline
    \makecell{The real intent\\ of users} & ``pink flowers'' & \makecell{``green and black \\ checkerboard pattern''}\\
    \hline
    \thead{Microsoft\\ Vision Studio} & \thead{``a pink  circle on \\ a brown surface''} & \thead{``a white couch with\\ black and green stripes''} \\
    \hline
    \thead{RAM\\ + GSAM} & \thead{``heart'', ``pink'',\\ ... ...} & \thead{``green'', ``pillow'',\\ ``couch'', ... ...} \\
    \hline
    MagicQuill & ``lightpink, marble'' & \makecell{``black, forestgreen, \\stone, fireplace''}\\
    \hline
    \thead{MLLM\\ (Ours)} & \thead{``Delicate pink\\ blossoms\\ scattered across\\ mountain slopes''} & \thead{``A Scottish tartan\\ fabric with alternating\\ black and green\\ checkered patterns''} \\
    \hline
  \end{tabular}
\end{table}

Furthermore, since this is an ill-posed problem, it is true that a single MLLM prediction cannot always be correct. However, our experiments show that its accuracy reaches nearly 80\%, significantly outperforming other methods. Moreover, by adjusting the number of predictions, it can cover the range of human imagination. When the model predicts four editing intents simultaneously, the probability of including the correct one increases to 85.1\%.

\begin{figure}[htbp]
  \centering
   \includegraphics[width=0.9\linewidth]{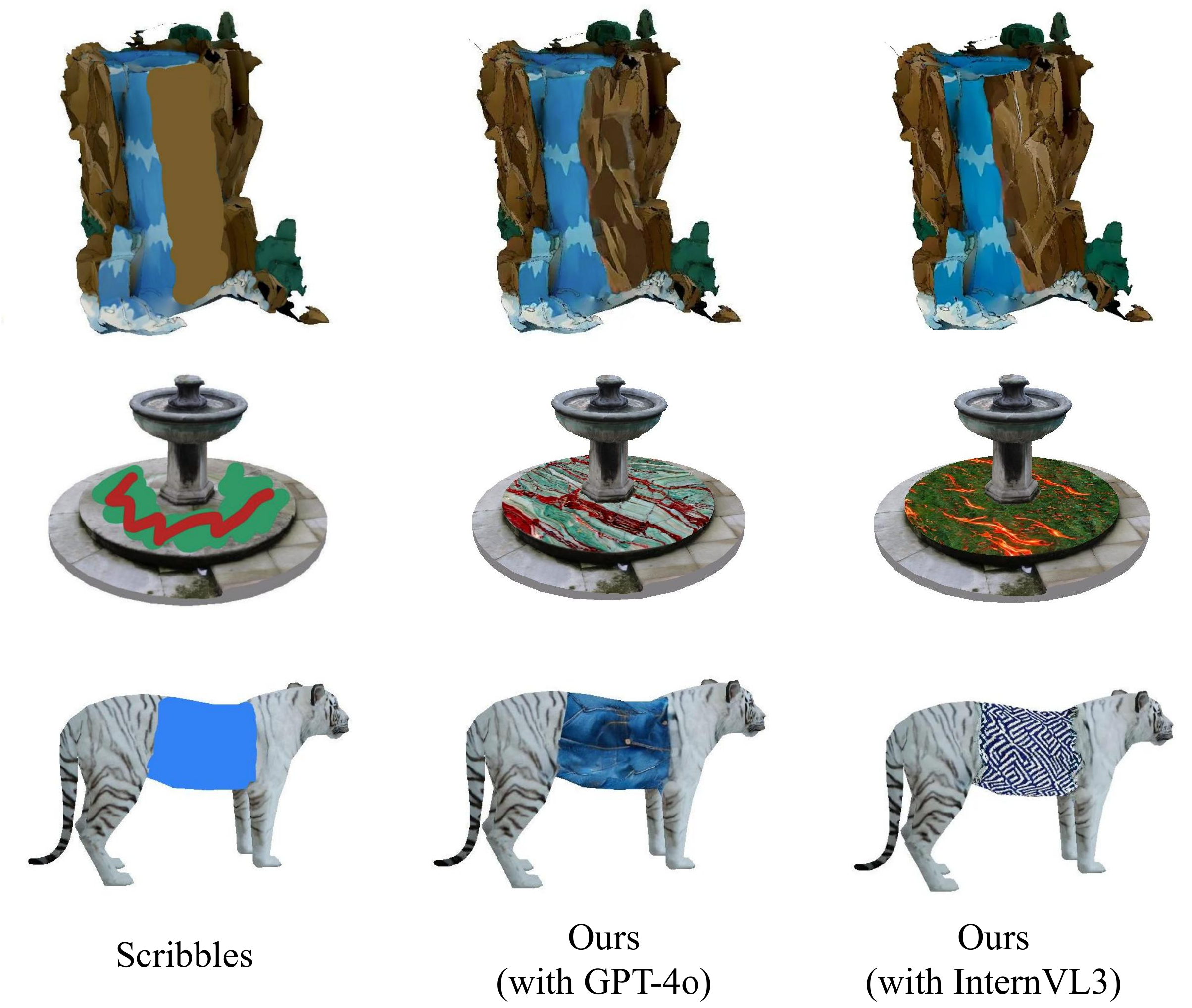}
   \caption{Qualitative results with different MLLMs.}
   \label{fig:mllmcmp}
\end{figure}

\begin{figure}[htbp]
  \centering
   \includegraphics[width=1.0\linewidth]{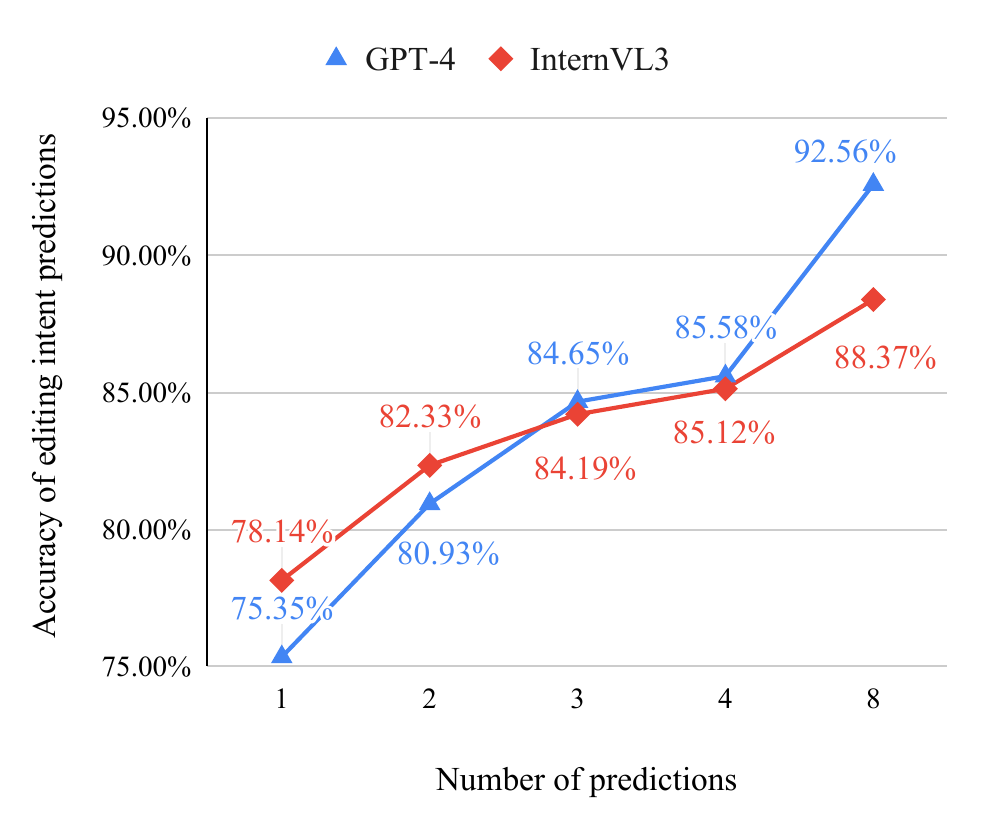}
   \caption{Relationship between prediction accuracy and the number of predictions across different MLLMs.}
   \label{fig:accuracy}
\end{figure}

{\bf MLLM Selection.} 
We have demonstrated the compatibility of MLLMs with our task. Leveraging this capability, our method can address editing tasks that are challenging for traditional approaches. Notably, ScribbleSense is not tied to any specific MLLM and does not require carefully engineered prompts. In this section, we evaluate the generalizability of our method by testing it with different MLLMs. 

Besides the commonly used GPT-4, we also compare with the latest open-source model, InternVL3. As shown in Figure \ref{fig:mllmcmp}, although the two models sometimes produce different intent predictions, both generate textures that are detailed and semantically coherent.

We further compare their prediction accuracy under varying numbers of generated predictions, a tunable parameter in our method. As illustrated in Figure \ref{fig:accuracy}, both models achieve comparable and satisfactory accuracy, which improves as the number of predictions increases, demonstrating the strong potential of MLLMs for this task.

\subsection{Ablation Study}

\begin{figure*}[htbp]
  \centering
   \includegraphics[width=0.95\linewidth]{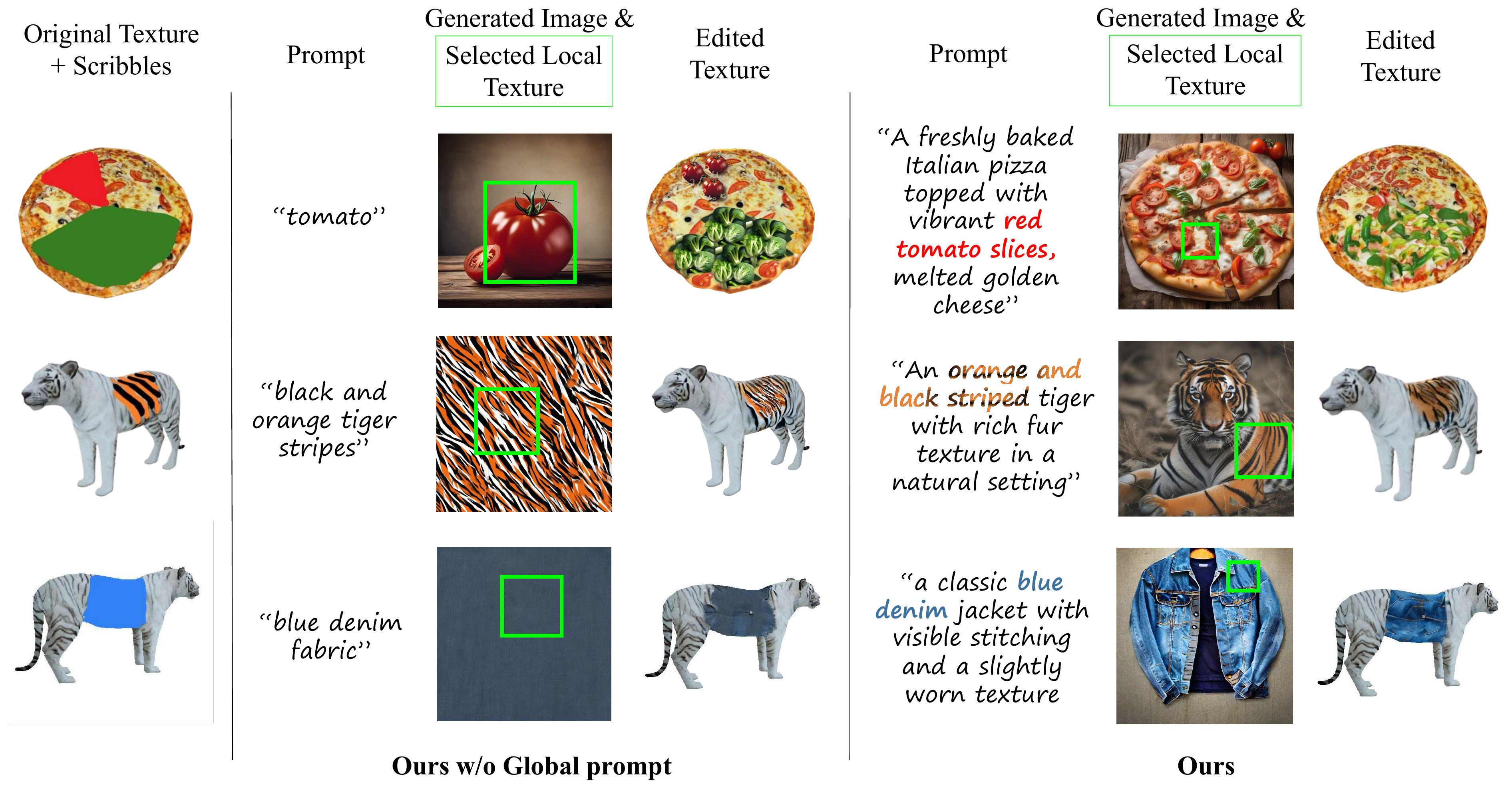}
   \caption{Ablation results of the global prompt generation module.}
   \label{fig:ablation}
\end{figure*}

To evaluate the effectiveness of each module, we conducted ablation studies on the multi-view input, global prompt generation, and scribble area refinement modules.

After obtaining the user scribble input, we incorporate multi-view image information to help the LLM acquire more comprehensive model semantic information, enabling more accurate prediction of the editing intent. We first validate the contribution of multi-view information to prediction accuracy. As shown in Table \ref{tab:llmeval}, without incorporating multi-view information, the prediction accuracy drops noticeably compared to our full method.

After the LLM initially predicts the local semantics represented by the scribble, we generate an optimized global prompt to obtain higher-quality local textures. We also validate the effectiveness of this global prompt generation step.

For the group without the global prompt, we directly use the initially predicted local semantic prompt to generate textures, which are then passed through the patch selection module for texture filling.
As shown in Figure \ref{fig:ablation}, removing the global prompt degrades both local and final texture quality. In the first row, unoptimized local textures fail to preserve correct shapes, producing a nonsensical tomato pizza. In the second and third rows, the lack of global prompt optimization results in poor style and texture control, leading to lower-quality outputs.

In addition, to further assess the impact of the scribble area refinement module, we conduct additional ablation experiments under challenging cases where the scribbles are rough or incomplete. We visualize the improvements using multi-view results to show how the refinement module improves the reasonableness of the edited region.
\begin{table}[htbp]
\caption{Quantitative results of the ablation study.\label{tab:ablation_tab}}
\centering
\begin{tabular}{|c|c|c|}
\hline
Method & CLIP-Score$\uparrow$ & User Preference$\uparrow$ \\
\hline
Ours w/o global prompt & 30.67 & 24.19 \% \\
\hline
Ours w/o area refinement & 31.35 & 30.23 \% \\
\hline
 Ours & {\bf 31.90} & {\bf 45.58 \%} \\
\hline
\end{tabular}
\end{table}
\begin{figure}[htbp]
  \centering
   \includegraphics[width=0.9\linewidth]{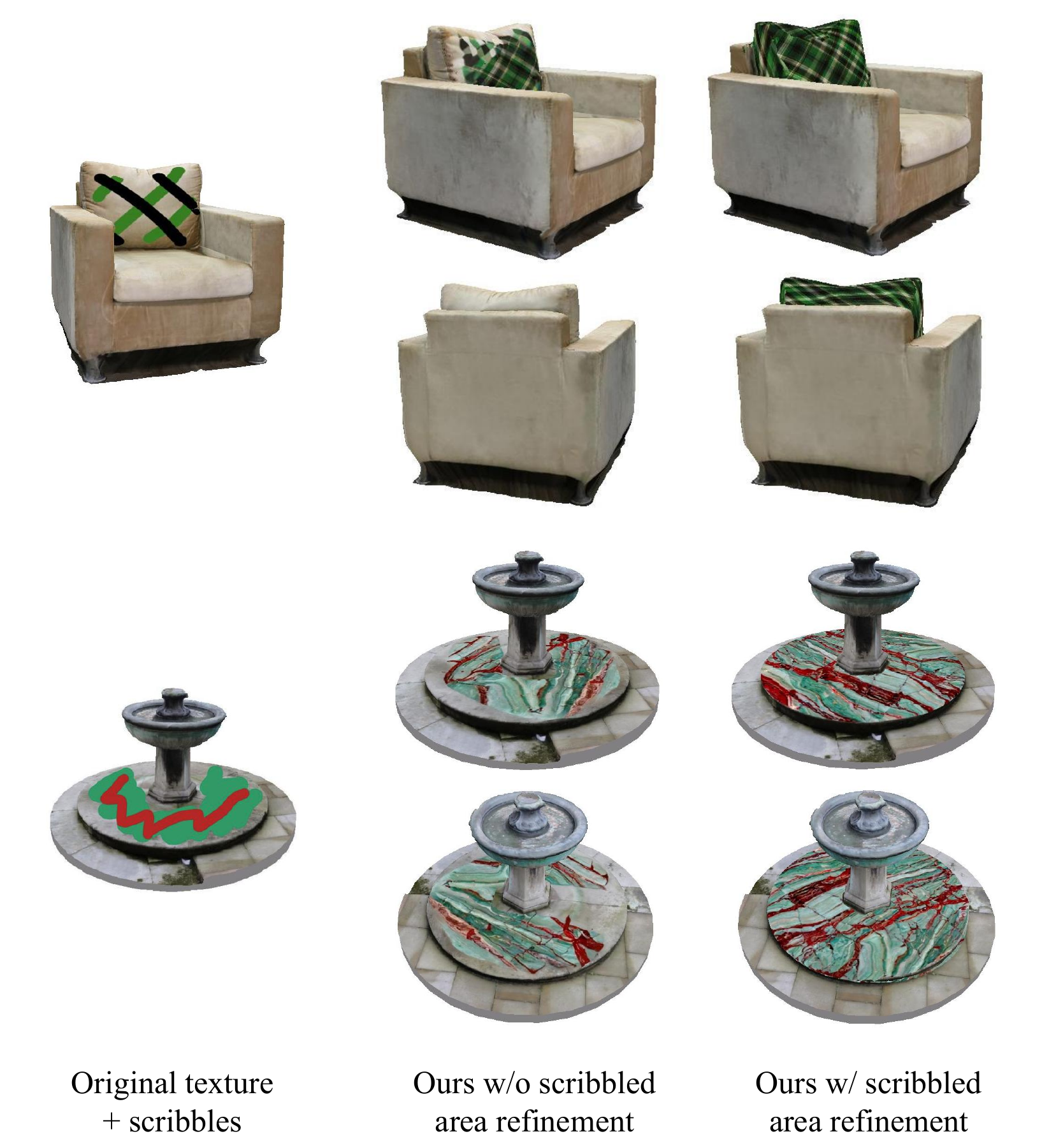}
   \caption{Comparison of results with and without the area refinement module under rough scribble inputs. For each example, the two rows present rendered images of the editing results from two different viewpoints.}
   \label{fig:ablationgeo}
\end{figure}

In the group without this module, the selected local texture patch is directly applied to the original scribbled area without any refinement. Figure \ref{fig:ablationgeo} presents a comparison between the results with and without the refinement module. It can be observed that, without the refinement step, the editing area is susceptible to boundary errors caused by freehand scribbles, resulting in unnatural appearances and a decline in overall quality. In contrast, with the refinement module incorporated, our method can more accurately interpret the user's intent and produce editing results that are more reasonable and better aligned with the underlying geometry. Notably, the area refinement module is not mandatory, and users can disable it if they wish the scribble to be followed strictly as drawn.

The quantitative results in Table \ref{tab:ablation_tab} further demonstrate that both the global prompt generation module and the scribble area refinement module contribute significantly to the overall performance.

\subsection{Exploration of Diverse Interactive Editing}
\label{sec:exploration}

\begin{figure*}[htbp]
  \centering
   \includegraphics[width=0.9\linewidth]{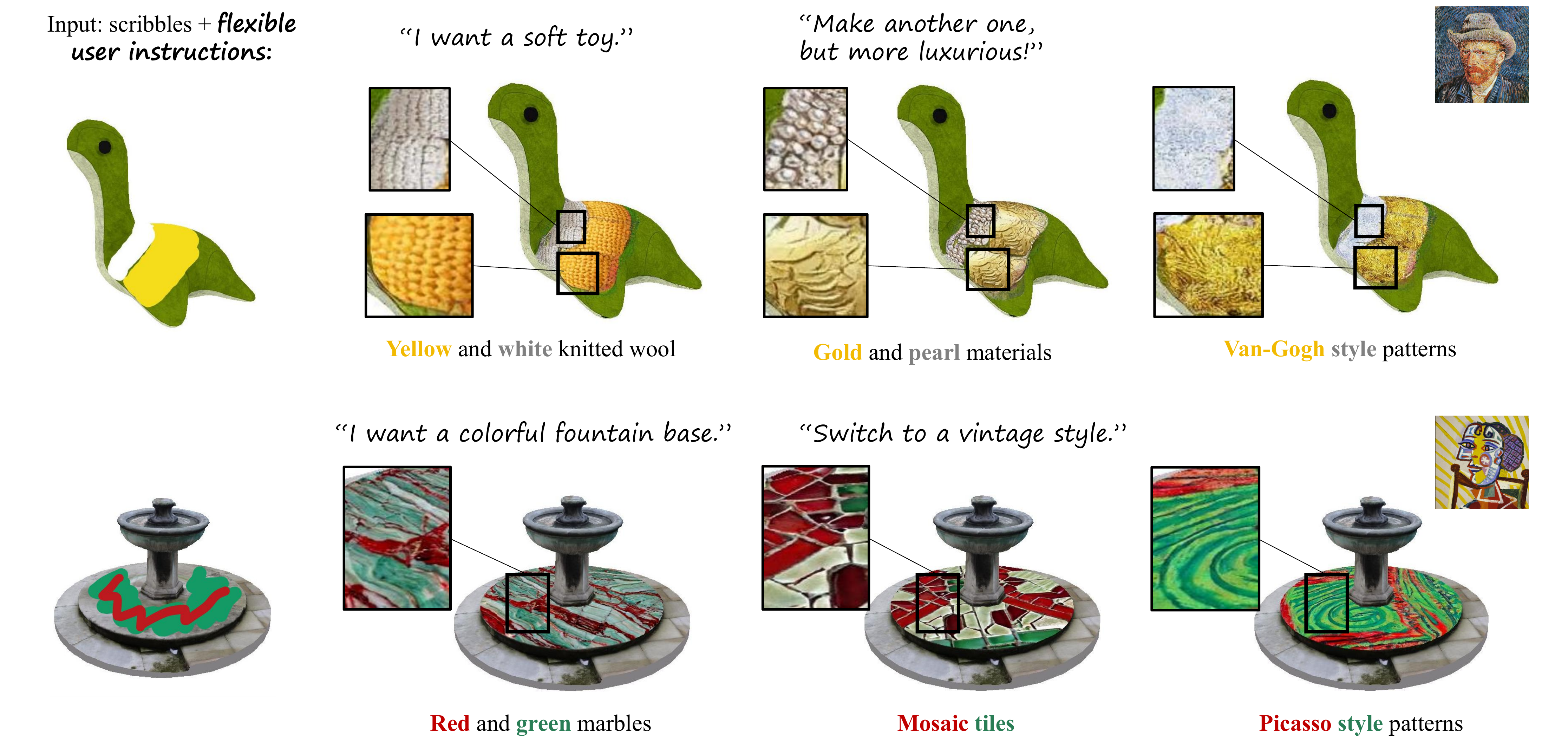}

   \caption{Diverse explorations of our method, including flexible prompt instructions and integration with stylized image generation models.}
   \label{fig:explore}
\end{figure*}

As mentioned earlier, our model integrates  MLLMs with image generation models. In previous experiments, we investigated the effectiveness of two widely adopted models, GPT-4 and SDXL, utilizing the most fundamental interaction manner. However, the capabilities of our method extend well beyond these initial configurations. In this section, we explore several advanced applications, providing insights for future model expansions.

LLMs excel in understanding natural language and responding to flexible user instructions. This capability enables us to offer more personalized prompts when predicting the semantic intent behind scribbles, thereby assisting the model in generating textures that align more closely with user expectations. As shown in Figure \ref{fig:explore}, when given the instruction ``more luxurious'', our model predicts the white and yellow scribbles as pearl and gold materials, respectively, and generates the final texture accordingly.

Moreover, our method is inherently plug-and-play, allowing for local texture generation that is not restricted to the capabilities of standard SD models. Instead, it can leverage different expert models designed for image synthesis. For example, models such as StyleDrop \cite{sohn2023styledrop}, ZipLoRA \cite{shah2024ziplora}, and InstantStyle \cite{wang2024instantstyle} can take a reference image with a prompt to generate textures in a consistent style. In Figure \ref{fig:explore}, we integrate StyleDrop into our method using a Van Gogh painting as a reference, so the generated textures inherit its artistic style. Similarly, users can flexibly incorporate other image generation models into our framework to produce textures tailored to their specific needs.

\begin{figure}[htbp]
  \centering
   \includegraphics[width=0.9\linewidth]{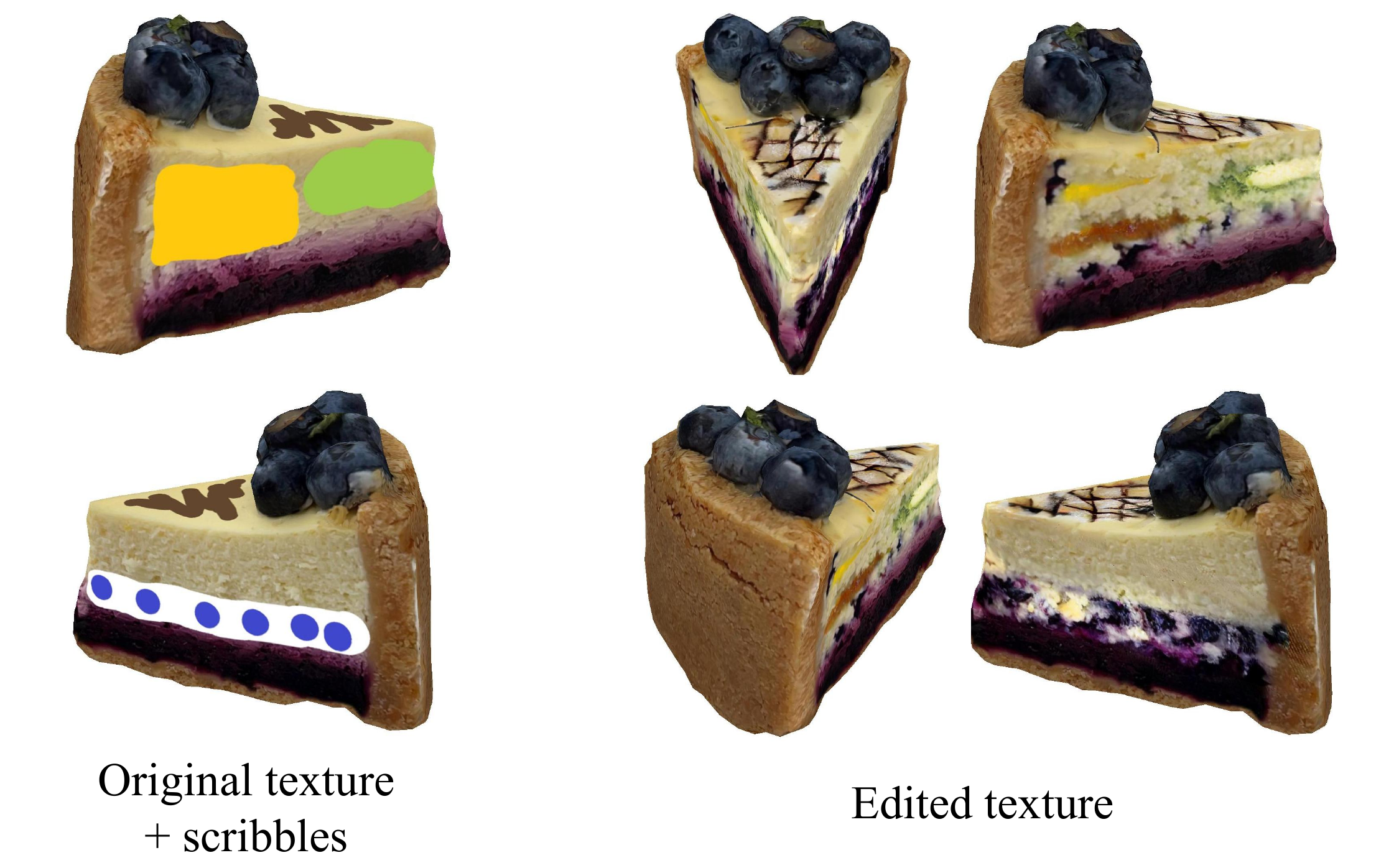}
   \caption{Results of simultaneously editing multiple regions using our method.}
   \label{fig:multiple_edit}
\end{figure}

Beyond diverse interaction modes, we also explore simultaneous multi-region edits. As shown in Figure \ref{fig:multiple_edit}, the model produces new textures that are both semantically consistent and visually natural.

\subsection{Limitation and Discussion}
\label{sec:limit}
\begin{figure}[htbp]
  \centering
   \includegraphics[width=0.95\linewidth]{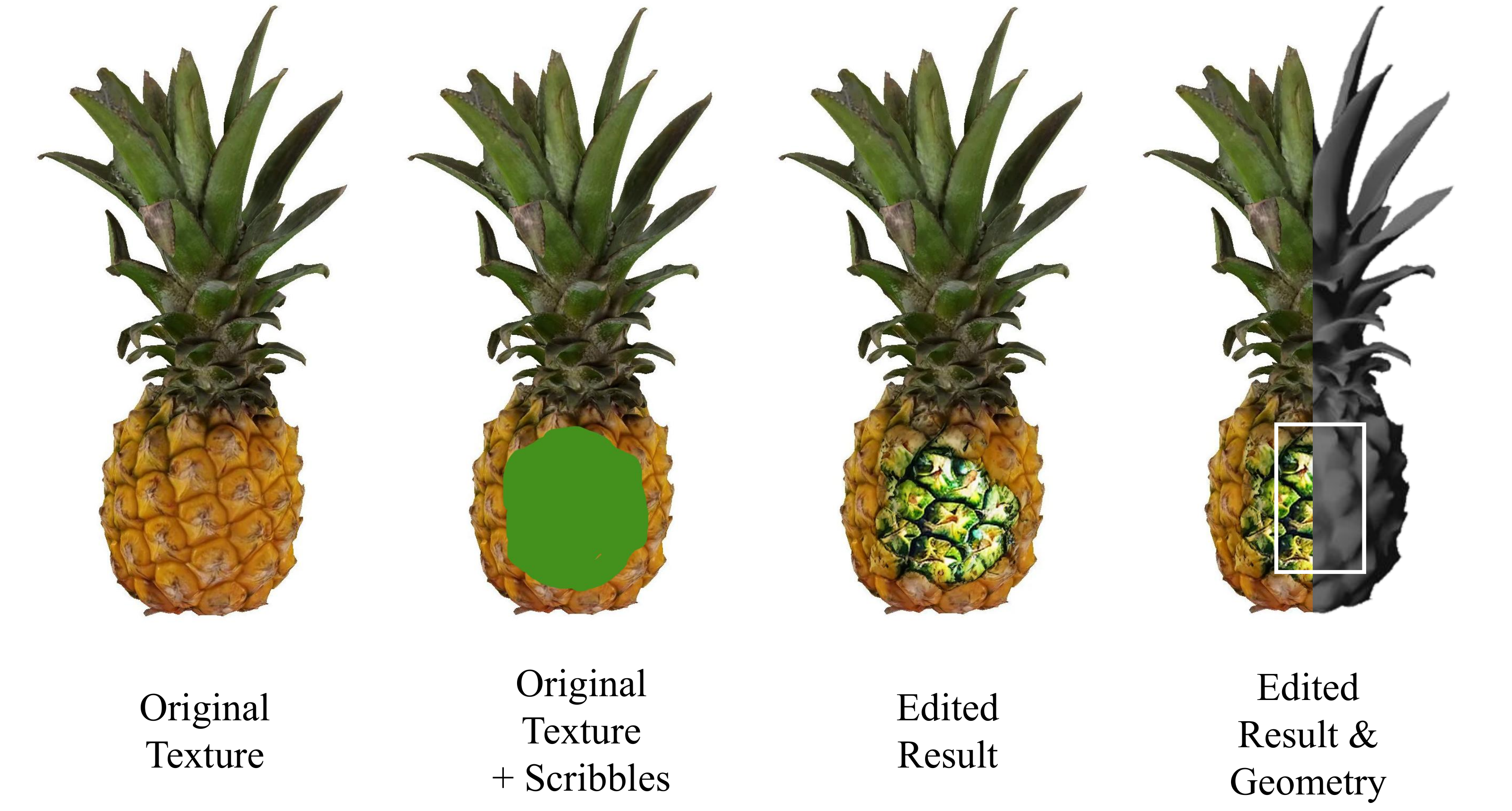}
   \caption{Failure case. When the model geometry is highly detailed, our method sometimes generates textures that do not align with the geometry.}
   \label{fig:failure_case}
\end{figure}

While our method demonstrates superior performance across multiple categories of examples, there remain aspects that have not yet been explored. For instance, the generated local textures rely on the color of the scribbles and the semantic cues of the original 3D model, without incorporating local geometric details. This leads to imperfect alignment between the edited textures and the underlying geometry in complex geometric scenes, as illustrated in Figure \ref{fig:failure_case}. Moreover, after generating local texture patches, we place them into the texture space, which makes our method more limited in cases where certain UV islands are excessively fragmented.

Since our method is designed to be plug-and-play, replacing certain components can endow it with different capabilities. In the scribble area refinement module, we currently adopt the SAM model, which achieves strong performance at present. Moreover, with the continued advancement of MLLMs, we believe our method has the potential to evolve into a more unified and general framework. In addition, incorporating depth-based image generation models into the local texture generation process could further improve the alignment between textures and geometry in complex geometric scenes.

\section{Conclusion}
We propose a scribble-based texture editing method, termed ScribbleSense, which leverages multimodal large language models to infer the editing intent behind scribbles, facilitating intuitive editing without the need for textual prompts. Our method converts scribbles into semantic information, refines this information into high-quality prompts using MLLMs, and synthesizes a cohesive global image. Finally, an inpainting model seamlessly integrates the local textures into the designated scribbled regions. Extensive qualitative and quantitative experiments demonstrate the effectiveness and potential of our method, as well as the compatibility of large language models with this task.

In future work, we plan to extend the application of vision-language models to a broader range of 3D scene generation and editing tasks, including more fine-grained sketch-based editing. Furthermore, we intend to optimize editing efficiency to enhance the interactive editing experience.




\bibliographystyle{ieeetr}
\bibliography{ref}

\vfill

\end{document}